\documentclass{bmvc2k}
\usepackage{times}
\usepackage{epsfig}
\usepackage{graphicx}
\usepackage{amsmath}
\usepackage{adjustbox}
\usepackage{booktabs}
\usepackage{tabularx}
\usepackage{enumitem}

\title{Taming the Tail: Leveraging Asymmetric Loss and Padé Approximation to Overcome Medical Image Long-Tailed Class Imbalance}

\addauthor{Pankhi Kashyap}{pankhikashyap.research@gmail.com}{1}
\addauthor{Pavni Tandon}{19d070044@iitb.ac.in}{1}
\addauthor{Sunny Gupta}{sunnygupta@iitb.ac.in}{1}
\addauthor{Abhishek Tiwari}{abhishektiwari.bm@gmail.com}{1}
\addauthor{Ritwik Kulkarni}{ritwik@oraiclebio.com}{2,3}
\addauthor{Kshitij Sharad Jadhav}{ kshitij.jadhav@iitb.ac.in}{1}

\addinstitution{
 Indian Institute of Technology, Bombay\\
 Mumbai, India
}
\addinstitution{University of Helsinki}
\addinstitution{Oraicle Biosciences LTD}
\runninghead{KASHYAP \MakeLowercase{et al.}}{TAMING THE TAIL}

\begin{document}

\maketitle

\begin{abstract}
Long-tailed problems in healthcare emerge from data imbalance due to variability in the prevalence and representation of different medical conditions, warranting the requirement of precise and dependable classification methods. Traditional loss functions such as cross-entropy and binary cross-entropy are often inadequate due to their inability to address the imbalances between the classes with high representation and the classes with low representation found in medical image datasets. We introduce a novel polynomial loss function based on Padé approximation, designed specifically to overcome the challenges associated with long-tailed classification. This approach incorporates asymmetric sampling techniques to better classify under-represented classes. We conducted extensive evaluations on three publicly available medical datasets and a proprietary medical dataset. Our implementation of the proposed loss function is open-sourced in the public repository: \href{https://github.com/ipankhi/ALPA}{https://github.com/ipankhi/ALPA}.

\end{abstract}

\section{Introduction}
\label{sec:intro}
Medical image classification is a crucial component in the development of effective diagnostic as well as prognostic tools \cite{9103969}. The utility of these tools often relies on the ability to manage and interpret large volumes of medical imaging data. However, a pervasive challenge encountered in these datasets is the prevalence of a long-tailed distribution—a scenario where the majority of data samples belong to a few dominant classes, while the remaining classes have significantly fewer samples \cite{liu2019largescale}. This imbalance poses significant challenges in training accurate classifiers, as conventional machine learning algorithms often struggle to learn from classes with limited samples \cite{krawczyk2016learning}. The existence of long tails in medical image datasets can be attributed to several factors, such as the rarity of certain medical conditions or diseases leading to a limited number of samples available for those classes \cite{obukhov2021proceedings}. As a result, these classes have few positive examples, making them challenging to detect and classify accurately. Furthermore, data collection in medical imaging is often biased towards common and easily accessible conditions, resulting in an uneven representation of different classes \cite{dahal2022hybrid}, \cite{Iqbal2023DynamicLF}. The challenges posed by long-tailed class distributions in medical image classification have thus prompted researchers to explore various solutions.

In their survey "Deep Long-Tailed Learning," Zhang \textit{et al} grouped existing solutions into three main categories: class re-balancing, information augmentation, and module improvement \cite{zhang2023deep}. These were further classified into nine sub-categories; 
Re-sampling methods, such as oversampling and undersampling, involve altering the class distribution in the training set \cite{Chawla_2002}, \cite{4717268}. Class-sensitive learning methods, like re-weighting \cite{lin2018focal}, \cite{48366} ,\cite{tan2020equalization} and re-margining \cite{cao2019learning} aim to re-balance training loss values for different classes promoting equitable learning, while logit adjustment techniques \cite{menon2021longtail} aim to re-calibrate the output probabilities of the classifier to account for the imbalanced class distribution. Transfer learning aims to enhance model training on a target domain by transferring knowledge from a source domain \cite{hendrycks2019using} and data augmentation techniques diversify datasets by either applying transformations directly to existing data or by utilizing generative AI methods, such as Generative Adversarial Networks (GANs) and Diffusion models, to create new samples \cite{perez2017effectiveness}, \cite{jimaging9030069}, \cite{rguibi2023improving}. Representation learning methods aim to learn more discriminative feature representations that can better separate different classes \cite{khosla2021supervised}, \cite{yang2022proco}, while classifier design involves optimizing the architecture and parameters of the classifier to improve its performance on long-tailed datasets by transferring geometric structures from head classes to tail classes \cite{Liu_2021_ICCV}. Decoupled training techniques decouple the training of the classifier into two stages: a representation learning stage and a classifier learning stage \cite{kang2020decoupling}. Finally, ensemble learning methods combine multiple classifiers, each trained on different subsets of the data or with different techniques, to improve classification performance \cite{zhou2020bbn}, \cite{li2020overcoming}. 

Loss functions play a crucial role in guiding model training. Class-sensitive loss functions are designed to mitigate the adverse effects of class imbalance by adjusting the contribution of each class to the overall loss calculation. These loss functions aim to ensure that the model does not disproportionately prioritize majority classes over minority ones during training. By doing so, they help alleviate the challenges associated with skewed class distributions and improve the model's ability to generalize across all classes. Focal loss, introduced by \cite{lin2018focal}, is a classic strategy to mitigate long-tailedness in classification tasks by dynamically adjusting the weighting of different examples during training to focus more on hard-to-classify samples. Similarly, class-balanced loss \cite{48366} assigns weights to different classes inversely proportional to their frequencies. Asymmetric loss \cite{benbaruch2021asymmetric} and asymmetric polynomial loss \cite{huang2023asymmetric} are variants of loss functions designed to penalize misclassifications of minority classes more heavily than majority classes. 

\section{Our contribution}
 
Polynomial expansions allow for the modeling of higher-order interactions between variables that linear models typically miss, thus providing a more nuanced and detailed depiction of data behaviors. Additionally, this method can be particularly useful in healthcare image analysis domains where capturing non-linear patterns is essential for predicting outcomes with high accuracy. By incorporating polynomial terms, models can approximate a wider range of functions, thereby adapting more effectively to the underlying complexities of the dataset \cite{leng2022polyloss}.

The Padé approximation \cite{weisstein_pade} is a mathematical technique that approximates a function through a ratio of two polynomials rather than relying solely on polynomial expansions. In earlier works, learnable activation functions based on the Padé approximation have shown promising performance \cite{molina2020pade}, \cite{article}. This method is particularly effective in modeling functions with singularities and provides a more accurate approximation over certain intervals. By applying the Padé approximation to the BCE loss function, we aim to achieve a more precise representation of the loss landscape, enabling our model to adjust more effectively to the true distribution of training data. Asymmetric focusing addresses the imbalance between the positive and negative classes by applying different weights to the loss contributions of each class. This technique is crucial in long-tail scenarios, where the minority class requires greater emphasis to ensure sufficient model sensitivity towards less frequent conditions. 
    
 In our research, 
\begin{itemize}[leftmargin=*]
    \item We introduce a novel approach to address the challenge of long-tailed medical image classification by proposing a Padé expansion-based polynomial loss function.
    \item Furthermore, by implementing an asymmetric focus, this loss function demonstrates enhanced classification performance for under-represented classes compared with other loss function-driven techniques in long-tailed problems.
    \item  We rigorously tested the efficacy of our method (\textbf{A}symmetric \textbf{L}oss with \textbf{P}adé \textbf{A}pproximation [\textbf{ALPA}]) across three publicly available medical image datasets in addition to a proprietary medical image dataset. 
\end{itemize} 


\section{Related Work}


The development of loss functions tailored for imbalanced datasets has been a focal point of research. The standard cross entropy loss is a commonly used loss function for classification tasks, defined as:

\begin{equation}\label{eq:multi-CE}
    \begin{cases}
        L_{CE}^{+} = -\sum_{i=1}^{K} y_i \log(\hat{y}_i), \\
        L_{CE}^{-} = -\sum_{i=1}^{K} (1 - y_i) \log(1 - \hat{y}_i),
    \end{cases}
\end{equation}
where \( K \) is the number of classes, and \( y_i \) and \( \hat{y}_i \) represent the ground-truth and estimated probabilities for class \( i \) respectively. However, when dealing with imbalanced datasets, the cross entropy loss (Equation \ref{eq:multi-CE}) treats all class samples equally and does not consider the imbalanced distribution. Thus, it tends to prioritize majority classes, leading to suboptimal performance on minority classes. 
Lin \textit{et al} \cite{lin2018focal} proposed Focal Loss (Equation \ref{eq:FL}) as a modification, which dynamically adjusts the loss weights based on the predicted probabilities. This enables Focal Loss to down-weigh the loss assigned to well-classified examples and focus more on difficult-to-classify instances. It is formulated as follows:
\begin{equation}\label{eq:FL}
    \begin{cases}
        L_{Focal}^{+} = \alpha_{+} (1 - \hat{y})^{\gamma} \log(\hat{y}) \\
        L_{Focal}^{-} = \alpha_{-} \hat{y}^{\gamma} \log(1 - \hat{y})
    \end{cases}
\end{equation}

where \( \alpha_{+} \) and \( \alpha_{-} \) are the balancing factors for positive and negative losses, respectively, and \( \gamma \) is the focusing parameter. Notably, setting \(\gamma = 0\) yields the binary cross-entropy loss. However, Focal Loss uses the same focusing parameter \( \gamma \) for both positive and negative losses. This can lead to suboptimal performance, especially in scenarios where the tail classes require different treatment compared to the head classes.

The Asymmetric Loss \cite{benbaruch2021asymmetric} introduces an asymmetric weighting scheme to alleviate the weaknesses of the Focal Loss. Equation \ref{eq:ASL} assigns different focusing parameters for positive and negative losses, allowing for separate optimization of the training of positive and negative samples. It is defined as:
\begin{equation}\label{eq:ASL}
    \begin{cases}
        L_{ASL}^{+} = (1 - \hat{y})^{\gamma_{+}} \log(\hat{y}) \\
        L_{ASL}^{-} = \hat{y}^{\gamma_{-}} \log(1 - \hat{y})
    \end{cases}
\end{equation}
where \( \gamma_{+} \) and \( \gamma_{-} \) are the focusing parameters for positive and negative losses respectively.

The Class-Balanced (CB) Loss \cite{48366} is another technique aimed at mitigating the challenges posed by class imbalance in training datasets and is formulated as follows:
\begin{equation}
\label{eq:CB_loss}
    L_{CB} = -\frac{1}{K} \sum_{k=1}^{K} \frac{1 - \beta^{\gamma}}{1 - \beta} \cdot y_k^{\gamma} \cdot \log(\hat{y}_k)
\end{equation}
where \( \gamma \) is the focusing parameter and \( \beta \) is a hyperparameter controlling the balance between the effective number of samples for each class and the average effective number of samples. Unlike traditional loss functions, CB loss (Equation \ref{eq:CB_loss}) introduces a mechanism to dynamically adjust the weights of different classes during the training process. This adjustment is based on the effective number of samples for each class, thereby ensuring that minority classes receive higher weights compared to majority classes. In \cite{jamal2020rethinking} Jamal \textit{et al} shows that class-balanced loss can underperform due to the domain gap between head and tail classes.
Similarly, the Label-Distribution-Aware Margin (LDAM) Loss \cite{cao2019learning} is a loss function designed to enhance the discriminative power of deep neural networks by explicitly maximizing the margins between different classes.  Unlike traditional loss functions like cross-entropy, LDAM loss focuses on optimizing the margins between classes in the feature space, thereby promoting better class separation and improved generalization performance. However, negative eigenvalues can persist in the LDAM loss landscape for tail classes due to insufficient data representation, leading to directions of negative curvature \cite{rangwani2022escaping}, making it inefficient for achieving effective generalization on tail classes. 

\section{METHOD}
\subsection{Padé approximants for BCE loss}
The BCE loss can be decomposed into C-independent binary classification subproblems:
\begin{equation}
L_{\text{BCE}} = \frac{1}{C} \sum_{i=0}^{C} \left( y_i L^{+} + (1 - y_i) L^{-} \right), \quad y_i \in \{1, 0\}
\end{equation}
where $L^{+} = -\log(\hat{y_i})$ is for the positive class, and $L^{-} = -\log(1 - \hat{y_i})$ is for the negative class. Here, $\hat{y_i}$ is the prediction probability after the sigmoid function.
We first define $L_{\text{BCE}}$ in Padé approximant form. For positive classes where $y_i=1$, we set the polynomial expansion point to be $1$; for negative classes where $y_i=0$, we set the expansion point to $0$. 
Thus, Padé approximants for the positive and negative classes for a single sample are: 
    \begin{equation}\label{eq:PadéAPL}
\begin{aligned}
    L^{+}_{\text{Padé}} &= \frac{a_0 + \sum_{m=1}^{M} a_m \hat{y}^m}{1 + \sum_{n=1}^{N} b_n \hat{y}^n},\\
    L^{-}_{\text{Padé}} &= \frac{c_0 + \sum_{m=1}^{M} c_m (1 - \hat{y})^m}{1 + \sum_{n=1}^{N} d_n (1 - \hat{y})^n}
\end{aligned}
\end{equation}
        where $\hat{y}$ represents the prediction probability of a single sample, while $M$ and $N$ represent the orders of the numerator and denominator polynomials, respectively, and $a_0$, $a_m$, $b_n$, $c_0$, $c_m$, and $d_n$ are coefficients of Padé approximants. 

\subsection{Derivation of the coefficients}
The conventional Padé approximation of order $m/n$ tends to reproduce the Taylor expansion of order \cite{weisstein_pade} $m+n$, and the coefficients are found by setting 
\begin{equation}
    \frac{P(x)}{Q(x)} = A(x)
\end{equation}
where \(P(x)\) is a numerator polynomial of order \(m\), \(Q(x)\) is the denominator polynomial of order \(n\) of Padé approximant , and \(A(x)\) is the Taylor expansion of order \(m+n\). \\
In terms of the Taylor Series Expansion, $L^+$ and $L^-$ are:
\begin{equation}\label{eq:TaylorAPL}
\begin{cases}
  L^+_{\text{Taylor}} = \sum_{k=1}^{\infty} (-1)^{k+1} \frac{(\hat{y_i} - 1)^k}{k} \\
  L^-_{\text{Taylor}} = - \sum_{k=1}^{\infty} \frac{\hat{y_i}^k}{k}
\end{cases}
\end{equation}

In line with previous research that highlights the effectiveness of the first-degree polynomial \cite{leng2022polyloss}, we adopt the first-order Padé approximation for our loss function. This approach sets both the numerator's and the denominator's orders to one, and for deriving the coefficients, we equate them with the respective Taylor series expansion of the second order. \\
The first order of \ref{eq:PadéAPL} would be:
\begin{equation}\label{eq:PadéAPL_1} 
\begin{cases}
    L^+_{\text{Padé}} \approx \frac{a_0 + a_1\hat{y_i}}{1 + b_1 \hat{y_i}} \\
    L^-_{\text{Padé}} \approx \frac{c_0 + c_1 (1 - \hat{y_i})}{1 + d_1 (1 - \hat{y_i})} 
\end{cases}
\end{equation}

Expanding \ref{eq:TaylorAPL} up to second order:
\begin{equation}\label{eq:TaylorAPL_1}
\begin{cases}
 L^+_{\text{Taylor}} \approx (\hat{y_i} - 1) - \frac{1}{2} (\hat{y_i} - 1)^2 \\
 L^-_{\text{Taylor}} \approx - \hat{y_i} - \frac{1}{2} \hat{y_i}^2
 \end{cases}
\end{equation}

By equating \ref{eq:PadéAPL_1} and \ref{eq:TaylorAPL_1}, we obtain the values of the coefficients $a_0= -1.5$, $a_1 = 1.5$, and $b_1 = 0$ , and the coefficients $c_0= -1$, $c_1 = 1$, and $d_1 = 0$. 

\subsection{Addition of asymmetric focusing mechanism and balancing factors}
Allowing separate optimization of the positive and negative samples, we add balancing factors and asymmetric focusing mechanism from \ref{eq:ASL}. Our proposed asymmetric loss based on Padé approximation becomes,  
\begin{equation}\label{eq:Final_loss}
L_{\text{ALPA}} = \sum_{i=1}^{N} \left[ \alpha y_i (1 - \hat{y_i})^{\gamma_{\text{pos}}} L^+_{\text{Padé}} + \beta (1 - y_i) \hat{y_i}^{\gamma_{\text{neg}}} L^-_{\text{Padé}} \right] \cdot W_i
\end{equation}
where \(N\) is the number of labels, $\hat{y_i}$ is the predicted probability and \(y_i\) is the binary target label for the \(i\)-th sample, \(\alpha\) and \(\beta\) are balancing parameters, \(\gamma_{pos}\) and \(\gamma_{neg}\) are focusing parameters, \(L^+_{\text{Padé}}\) and \(L^-_{\text{Padé}}\) are the Padé Approximation forms for positive and negative predictions, respectively. \(W_i\) is the weight for the \(i\)-th sample, calculated as \((1 - pt_i)^{\gamma}\), where \(pt_i\) is the predicted probability adjusted for the target label such that $pt_i = y_i\hat{y_i}+ (1 - y_i)(1 - \hat{y_i})$, and  $\gamma$ is the summation of focusing parameters, with \(\gamma_{pos}\) applied for positive targets and \(\gamma_{neg}\) for negative targets. 

We studied the effects of hyperparameters \(\alpha\), \(\beta\), \(\gamma_{pos}\) and \(\gamma_{neg}\) on the loss function and evaluated the loss function using the best-performing combination of values on the datasets used in this study. 

\subsection{Gradient Analysis}
Gradients play a pivotal role in the training process, guiding the adjustments of network weights with respect to the input logit $z$. In this section, following the work of \cite{benbaruch2021asymmetric}, we provide a comprehensive analysis of the loss gradients of ALPA compared to established loss functions such as Cross Entropy, Focal Loss, and Asymmetric Loss. \\
For ALPA, we have $L^-_{\text{ALPA}} = (\hat{y}_i)^{\gamma_{\text{neg}} + 1}$, thus, the negative gradient equation for the ALPA function is given by:
\[
\frac{{dL^-_{\text{ALPA}}}}{{dz}} = \frac{dL^-_{\text{ALPA}}}{d\hat{y}_i} \cdot \frac{d\hat{y}_i}{dz} = (\hat{y}_i)^{\gamma_{\text{neg}}+1} \cdot (1 - \hat{y}_i) \cdot (\gamma_{\text{neg}} + 1)
\]

where \( \hat{y}_i = \frac{1}{{1 + e^{-z}}} \) represents the predicted probability for the input logit \( z \) and \( \gamma_{\text{neg}} \) is the focusing parameter for negative targets.

\begin{figure}[htbp]
    \centering
    \includegraphics[width=0.5\linewidth]{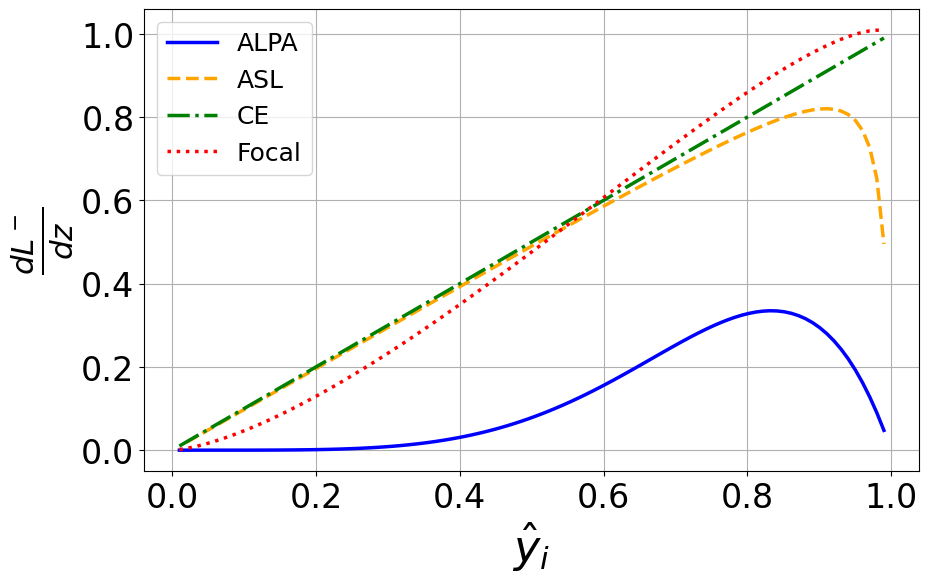}
    \caption{Comparison of loss gradients for ALPA ($\gamma_{\text{neg}}=4$), ASL ($m=0.01$, $\gamma_{\text{neg}}=0.01$), CE ($m=0$, $\gamma_{\text{neg}}=0$), and Focal Loss ($\gamma=0.5$)
}
    \label{fig:gradient_curve}
\end{figure}

The results of the gradient analysis are shown in Figure~\ref{fig:gradient_curve}.

We observe that the gradient for ALPA increases moderately as the probability $\hat{y}_i$ approaches 1. This suggests that our proposed loss function provides a consistent learning signal across the probability spectrum. It neither penalizes very harshly for misclassifications (when $\hat{y}_i$ is low) nor relaxes too much when the classification is correct (when $\hat{y}_i$) is high). Thus, ALPA appears to be a good choice for consistent learning across all probabilities. By focusing on harder examples and not over-penalizing the correctly classified ones, it achieves better generalization compared to other losses. 

\section{Experimental Setup}
\subsection{Datasets}\label{sec:Datasets}
The APTOS 2019 BD dataset \cite{bodapati2020blended} includes data from individuals diagnosed with varying levels of Diabetic Retinopathy (DR), categorized into five classes: No DR, Mild DR, Moderate DR, Severe DR, and Proliferative DR. The DermMNIST dataset \cite{yang2023medmnist} comprises 450x600 pixel images of various skin diseases classified into seven categories: Melanoma, Melanocytic Nevus, Basal Cell Carcinoma, Actinic Keratosis, Benign Keratosis, Dermatofibroma, and Vascular Lesion. The BoneMarrow dataset \cite{matek2021expert} contains expertly annotated cells from bone marrow smears of 945 patients, classified into 17 types including Basophil (BAS), Blast (BLA), Erythroblast (EBO), and more. The Oraiclebio dataset, which remains proprietary, includes 3,643 images of oral regions featuring 52 classes of precancerous and cancerous lesions.

Details of these datasets are summarized in Table \ref{tab:datasets}, where the imbalance ratio, defined as $N_{max}/N_{min}$ (with $N$ representing the sample count per class), illustrates the significance of the long-tailed distribution. For experimentation, each dataset was split 80-20 into training and testing sets, and a 5-fold cross-validation strategy was used during training to enhance model reliability.
\begin{table}[]
\begin{center}
\caption{Details of long-tailed medical datasets.}
\small
\begin{tabular}{c|c|c|c}
\hline
Dataset    & Classes & Samples & Imbalance Ratio \\ \hline
APTOS2019  & 5       & 3,662   & 10              \\ \hline
DermaMNIST & 7       & 10,015  & 58              \\ \hline
BoneMarrow & 17      & 147,904  & 621.79              \\ \hline
Oraiclebio    & 52      & 3643      & 164             \\ \hline
\end{tabular}
\label{tab:datasets}
\end{center}
\end{table}

\subsection{Implementation}
We use ConvNeXT-B \cite{liu2022convnet} as the backbone for the proposed loss. We resize the input images as $256$ x $256$ and exploit the data augmentation schemes following the previous work \cite{azizi2021big,chen2019multi}. We train our networks using the Adam optimizer with $0.9$ momentum and $0.001$ weight decay. The batch size is $128$, and the initial learning rate is set to $1 \mathrm{e}{-4}$. Our networks are trained on PyTorch version 2.2.1 with RTX A6000 GPUs. We use accuracy, balanced accuracy and F1-score as evaluation metrics for this study. 

\section{Results}
In this section, we present experimental results validating the effectiveness of the proposed \textbf{ALPA} function. We first analyze the impact of hyperparameters on the loss function and then compare \textbf{ALPA} with state-of-the-art loss functions like Asymmetric Loss, Focal Loss and Cross Entropy.

\subsection{Effect of the hyperparameters}
To evaluate the effect of hyperparameters, we experimented as follows:
\begin{itemize}
    \item \textbf{Loss v1}: Hyperparameters were randomly set as \(\alpha = 1\), \(\beta = 1\), \(\gamma_{pos} = 0\), and \(\gamma_{neg} = 4\). This is indicated as Loss v1 in Table \ref{tab:ablation}.
    \item \textbf{Loss v2 (\(L_{\text{ALPA}}\))}: Using random search, hyperparameters were optimized within the ranges \(\alpha\) and \(\beta\) (\(0.5\) to \(2\)), and \(\gamma_{pos}\) and \(\gamma_{neg}\) (\(0\) to \(5\)). Final values were \(\alpha = 0.875\), \(\beta = 1.625\), \(\gamma_{pos} = 0\), and \(\gamma_{neg} = 4\). This is indicated as Loss v2 in Table \ref{tab:ablation}.
    \item \textbf{Loss v3}: Incorporating Hill Loss \cite{zhang2021simple} following \cite{park2023robust}, we added \(\lambda - \hat{y}_i\) to \(L^-\) (\(\lambda = 1.5\)), optimizing via random search to \(\alpha = 1.25\), \(\beta = 2\), \(\gamma_{pos} = 3\), and \(\gamma_{neg} = 2\). This is indicated as Loss v3 in Table \ref{tab:ablation}.
\end{itemize}

Results on the APTOS2019 dataset for these settings are shown in Table \ref{tab:ablation}. We focused on detecting crucial cases like Proliferative DR and examined the performance of underrepresented classes to proceed with Loss v2. From here on, Loss v2 is referred to as \(L_{\text{ALPA}}\).

\begin{table}[ht]
\centering
\caption{Performance metrics for different versions of Loss.}
\label{tab:ablation}
\small
\begin{tabular}{l|c|cc|cc|ccc}
\hline
Classes       & Number of  & \multicolumn{2}{c|}{Loss v1} & \multicolumn{2}{c|}{Loss v2} & \multicolumn{2}{c}{Loss v3} \\ 
              &    training samples     & Acc         & F1-score     & Acc         & F1-score     & Acc         & F1-score     \\ \hline
No DR         & $1454$   & $98.86$       & $0.98$         & $98.01$       & $0.98$         & $98.86$       & $0.98$         \\
Mild          & $786$      & $64.71$       & $0.62$         & $58.82$       & $0.59$        & $26.47$       & $0.37$         \\
Moderate      & $302$     & $88.26$       & $0.76$         & $86.38$       & $0.79$         & $92.96$       & $0.79$         \\
Severe        & $230$     & $2.78$       & $0.5$         & $33.33$       & $0.44$        & $36.11$       & $0.39$         \\
Proliferative DR & $157$     & $23.08$       & $0.36$         & $47.69$       & $0.58$        & $30.77$       & $0.45$         \\
\hline
\end{tabular}
\end{table}

\subsection{Comparison with existing methods}
We compare our proposed loss function with state-of-the-art methods such as ASL, Focal Loss, LDAM, and CE on the datasets listed in Section \ref{sec:Datasets}. Results on the publicly available datasets are presented in Tables \ref{tab:APTOS_comparison}, \ref{tab:HAM_comparison}, and \ref{tab:BM_comparison}, while the results for LDAM loss functions can be found in the supplementary materials. \textbf{ALPA} consistently excels in classes with fewer samples while maintaining competitive accuracy in classes with higher representation. In terms of balanced accuracy, \textbf{ALPA} surpasses all other loss functions across the three public datasets. 

\begin{table}[ht]
\centering
\caption{Comparison of different loss functions on the APTOS2019 dataset.}
\label{tab:APTOS_comparison}
\resizebox{\textwidth}{!}{%
\begin{tabular}{l|c|cc|cc|cc|cc} 
\toprule
Classes & Number of & \multicolumn{2}{c|}{ALPA} & \multicolumn{2}{c|}{ASL} &\multicolumn{2}{c|}{CE} & \multicolumn{2}{c}{FOCAL} \\ 

 & training samples& Acc & f1-score & Acc & f1-score & Acc & f1-score & Acc & f1-score \\ 
\midrule
No DR & $1454$ & $98.01$ & $0.98$ &$98.58$ & $0.98$ & $98.86$ & $0.97$ & $\textbf{99.43}$ & $0.94$ \\ 
Mild & $786$ & $\textbf{58.82}$ & $0.59$ & $41.18$ & $0.50$ &$16.18$ & $0.27$ & $10.29$ & $0.16$  \\ 
Moderate & $302$ & $86.38$ & $0.79$ & $90.61$ & $0.77$ & $\textbf{97.18}$ & $0.76$ & $88.73$ & $0.75$ \\ 
Severe & $230$ & $\textbf{33.33}$ & $0.44$ & $11.11$ & $0.17$ &$8.33$ & $0.14$ & $8.33$ & $0.15$ \\ 
Proliferative DR & $157$ & $\textbf{47.69}$ & $0.58$ & $43.08$ & $0.57$ &$16.92$ & $0.28$ & $30.77$ & $0.41$ \\ 
\midrule
\multicolumn{2}{l|}{Balanced Accuracy} &  & $\textbf{0.65}$ &  & $0.57$ &  & $0.47$ &  & $0.48$ \\ 
\bottomrule
\end{tabular}%
}
\end{table}

\begin{table}[ht]
\centering
\caption{Comparison of different loss functions on the DermaMNIST dataset.}
\label{tab:HAM_comparison}
\resizebox{\textwidth}{!}{%
\begin{tabular}{l|c|cc|cc|cc|cc}
\toprule
Classes & Number of & \multicolumn{2}{c|}{ALPA} & \multicolumn{2}{c|}{ASL} &\multicolumn{2}{c|}{CE} & \multicolumn{2}{c}{FOCAL}\\  

 & training samples& Acc & f1-score & Acc & f1-score & Acc & f1-score & Acc & f1-score \\ 
\midrule
akiec   & $256$   & $4.23$  & $0.08$ & $\textbf{16.9}$&$0.29$ & $0.00$ & $0.00$ & $1.41$ & $0.03$ \\
bcc     & $406$  & $19.44$ & $0.31$  & $\textbf{26.85}$& $0.37$ &$5.56$ & $0.10$ & $\textbf{26.85}$ & $0.38$\\ 
bkl     & $882$  & $\textbf{35.94}$ & $0.45$  & $18.43$ & $0.28$ & $17.97$ & $0.26$ & $1.84$ & $0.04$ \\ 
df      & $88$   & $\textbf{33.33}$ & $0.39$  & $22.22$ & $0.35$ & $3.70$ & $0.07$ & $7.41$ & $0.10$ \\ 
mel     & $885$  & $\textbf{10.09}$ & $0.18$  & $8.33$ & $0.15$ & $0.44$ & $0.01$ & $1.32$ & $0.03$ \\ 
nv      & $5375$ & $98.50$ & $0.86$  & $96.62$ & $0.84$ & $\textbf{99.70}$ & $0.82$ & $98.27$ & $0.82$ \\ 
vasc    & $120$  & $\textbf{72.73}$ & $0.35$ & $54.55$ & $0.19$ & $36.36$ & $0.37$ & $81.82$ & $0.42$ \\ 
\midrule
\multicolumn{2}{l|}{Balanced Accuracy} &  & $\textbf{0.39}$ &  & $0.35$ &  & $0.23$ &  & $0.31$ \\ 
\bottomrule
\end{tabular}%
}
\end{table}

\begin{table}[ht]
\centering
\caption{Comparison of different loss functions on the Bone Marrow dataset.}
\label{tab:BM_comparison}
\resizebox{\textwidth}{!}{%
\begin{tabular}{l|c|cc|cc|cc|cc}
\toprule
Classes & Number of & \multicolumn{2}{c|}{ALPA} & \multicolumn{2}{c|}{ASL} & \multicolumn{2}{c|}{CE} & \multicolumn{2}{c}{FOCAL} \\ 
 & training samples & Acc & f1-score & Acc & f1-score & Acc & f1-score & Acc & f1-score \\ 
\midrule
BAS   & 348   & 54.84  & 0.68 & \textbf{55.91} & 0.68 & 46.24 & 0.61  & 51.61 & 0.66 \\ 
BLA   & 9569  & 85.94  & 0.87 & 87.06 & 0.88 & \textbf{89.93} & 0.88  & 87.52 & 0.88 \\ 
EBO   & 21883 & 95.94  & 0.96 & 96.23 & 0.96 & 96.21 & 0.96  & \textbf{96.34} & 0.96 \\ 
EOS   & 4719  & \textbf{97.34} & 0.97 & 97.25 & 0.96 & \textbf{97.34} & 0.97  & 96.99 & 0.97 \\ 
FGC   & 41    & \textbf{83.33} & 0.77 & 66.67 & 0.73 & \textbf{83.33} & 0.77  & 50.00 & 0.60 \\ 
HAC   & 339   & 58.57  & 0.71 & 67.14 & 0.77 & \textbf{72.86} & 0.81  & 71.43 & 0.80 \\
KSC   & 38    & \textbf{100.00} & 1.00 & 75.00 & 0.75 & 75.00 & 0.86  & \textbf{100.00} & 0.89 \\ 
LYI   & 54    & 36.36 & 0.40 & \textbf{36.98} & 0.44 & 27.77 & 0.38  & 9.09 & 0.14 \\ 
LYT   & 20911 & 94.13 & 0.94 & \textbf{94.92} & 0.94 & 94.75 & 0.94  & 94.62 & 0.94 \\ 
MMZ   & 2479  & 39.58 & 0.46 & 36.98 & 0.45 & 56.60 & 0.52  & \textbf{58.33} & 0.56 \\ 
MON   & 3230  & 74.81 & 0.78 & \textbf{79.38} & 0.79 & 76.91 & 0.77  & 79.01 & 0.78 \\ 
MYB   & 5238  & 68.76 & 0.70 & 70.20 & 0.70 & 61.26 & 0.68  & \textbf{73.09} & 0.73 \\ 
NGB   & 7967  & 67.82 & 0.69 & 69.42 & 0.71 & 66.12 & 0.71  & \textbf{72.06} & 0.74 \\ 
NGS   & 23628 & \textbf{94.24} & 0.91 & 92.86 & 0.92 & 93.32 & 0.92  & 93.65 & 0.93 \\ 
PEB   & 2196  & 75.55 & 0.75 & 77.76 & 0.78 & 73.71 & 0.76  & \textbf{78.68} & 0.78 \\ 
PLM   & 6137  & \textbf{93.90} & 0.93 & 91.49 & 0.93 & 92.16 & 0.93  & 91.69 & 0.93 \\ 
PMO   & 9546  & 85.58 & 0.86 & \textbf{88.40} & 0.86 & 88.19 & 0.86  & 84.03 & 0.85 \\ 
\midrule
\multicolumn{2}{l|}{Balanced Accuracy} &  & \textbf{0.77} &  & 0.75 &  & 0.76  &  & 0.76 \\ 
\bottomrule
\end{tabular}%
}
\end{table}

In the Oraiclebio dataset, \textbf{ALPA} delivers the highest accuracy in 23 classes, achieving a balanced accuracy of 51.06\% and performing similarly to ASL, which attains a balanced accuracy of 52\%, demonstrating its robustness across varying data support levels. Meanwhile, CE averages 47.92\% accuracy, and Focal Loss performs significantly worse, with an average balanced accuracy of just 22\%. Thus, \textbf{ALPA} stands out for its ability to handle diverse and imbalanced datasets effectively.

\section{Conclusion}
In this study, we present a Padé approximation-based loss function with asymmetric focusing, tailored for multi-class classification tasks with long-tailed distributions. Our proposed loss function demonstrates competitive and superior performance on long-tailed datasets when benchmarked against previous state-of-the-art approaches. We believe that our findings can serve as a valuable resource for future research, offering a foundation for further development and integration into new studies.

\section{Future work}
The learning process of the model is intrinsically tied to data representation. Modifying the loss function alone, however, may have limited potential for performance improvement. To enhance class-wise accuracy, integrating loss functions with data augmentation strategies and data generation pipelines presents a promising approach. Data augmentation artificially expands the training dataset by creating modified versions of existing images, while data generation pipelines synthesize entirely new samples. These methods can help balance class representation and bolster model robustness. A more thorough exploration of these techniques in future work could offer substantial benefits in addressing class imbalance.

\section*{Acknowledgements}
This work was supported through the funded project by Oraicle Biosciences LTD (OraiBio) to Indian Institute of Technology, Bombay.


\bibliography{egbib}

\end{document}


\maketitle 

In this supplementary document, we present the experimental results obtained using LDAM loss and BCE loss. Additionally, we include several visual representations of the proprietary dataset utilized in this study. 

\begin{figure}[h]
    \centering
    \includegraphics[width=0.7\linewidth]{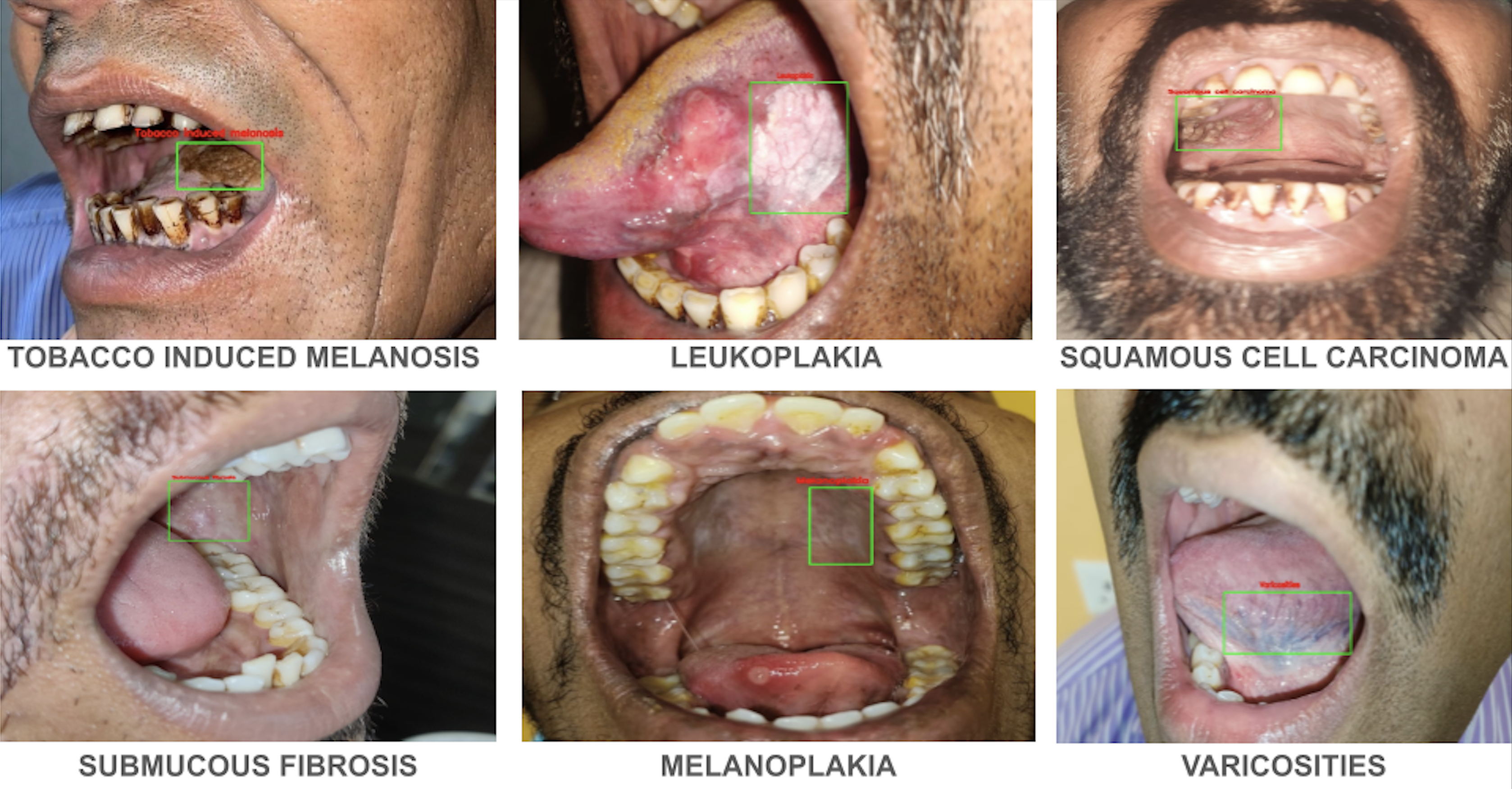}
    \caption{Images from a few classes of Oraiclebio 
    dataset}
    \label{fig:oraicle_images}
\end{figure}

Figure\ref{fig:oraicle_images} displays images from several classes within the Oraiclebio dataset. The green boxes highlight the regions containing lesions.

\clearpage
Tables \ref{tab:LDAM_comparison} and \ref{tab:BCE_comparison} present a comparison of classification accuracies using LDAM loss and BCE loss, respectively, across different classes from the public medical datasets analyzed in this study.
\begin{table}[h!]
\centering
\caption{Results of LDAM loss function across different datasets}
\label{tab:LDAM_comparison}
\begin{adjustbox}{max width=\textwidth}
\begin{tabular}{l|l|c|cc}
\toprule
Dataset & Classes & Training Samples & Acc & f1-score \\ 
\midrule
\multirow{5}{*}{APTOS2019} 
& No DR & 1454 & 13.39 & 0.15 \\
& Mild & 786 & 36.76 & 0.10 \\
& Moderate & 302 & 0.00 & 0.00 \\
& Severe & 230 & 0.00 & 0.00 \\
& Proliferative DR & 157 & 0.00 & 0.00 \\
\midrule
\multirow{7}{*}{DermaMNIST} 
& akiec & 256 & 100.00 & 0.07 \\
& bcc & 406 & 4.63 & 0.07 \\
& bkl & 882 & 0.00 & 0.00 \\
& df & 88 & 0.00 & 0.00 \\
& mel & 885 & 0.00 & 0.00 \\
& nv & 5375 & 0.00 & 0.00 \\
& vasc & 120 & 0.00 & 0.00 \\
\midrule
\multirow{17}{*}{Bone Marrow} 
& BAS & 348 & 1.08 & 0.01 \\
& BLA & 9569 & 99.96 & 0.15 \\
& EBO & 21883 & 0.00 & 0.00 \\
& EOS & 4719 & 0.00 & 0.00 \\
& FGC & 41 & 0.00 & 0.00 \\
& HAC & 339 & 0.00 & 0.00 \\
& KSC & 38 & 0.00 & 0.00 \\
& LYI & 54 & 0.00 & 0.00 \\
& LYT & 20911 & 0.00 & 0.00 \\
& MMZ & 2479 & 0.00 & 0.00 \\
& MON & 3230 & 0.00 & 0.00 \\
& MYB & 5238 & 0.00 & 0.00 \\
& NGB & 7967 & 0.00 & 0.00 \\
& NGS & 23628 & 0.00 & 0.00 \\
& PEB & 2196 & 0.00 & 0.00 \\
& PLM & 6137 & 0.00 & 0.00 \\
& PMO & 9546 & 0.00 & 0.00 \\
\bottomrule
\end{tabular}
\end{adjustbox}
\end{table}

\begin{table}[h!]
\centering
\caption{Results of BCE loss function across different datasets}
\label{tab:BCE_comparison}
\begin{adjustbox}{max width=\textwidth}
\begin{tabular}{l|l|c|cc}
\toprule
Dataset & Classes & Training Samples & Acc & f1-score \\ 
\midrule
\multirow{5}{*}{APTOS2019} 
& No DR & 1454 & 99.43 & 0.91 \\
& Mild & 786 & 11.76 & 0.17 \\
& Moderate & 302 & 84.04 & 0.75 \\
& Severe & 230 & 13.89 & 0.23 \\
& Proliferative DR & 157 & 16.92 & 0.28 \\
\midrule
\multirow{7}{*}{DermaMNIST} 
& akiec & 256 & 4.23 & 0.08 \\
& bcc & 406 & 7.86 & 0.18 \\
& bkl & 882 & 22.04 & 0.43 \\
& df & 88 & 26.76 & 0.35 \\
& mel & 885 & 6.19 & 0.10 \\
& nv & 5375 & 93.16 & 0.89 \\
& vasc & 120 & 70.00 & 0.28 \\
\midrule
\multirow{17}{*}{Bone Marrow} 
& BAS & 348 & 51.61 & 0.65 \\
& BLA & 9569 & 89.39 & 0.88 \\
& EBO & 21883 & 96.06 & 0.96 \\
& EOS & 4719 & 97.34 & 0.97 \\
& FGC & 41 & 66.67 & 0.73 \\
& HAC & 339 & 64.29 & 0.76 \\
& KSC & 38 & 25.00 & 0.40 \\
& LYI & 54 & 9.09 & 0.15 \\
& LYT & 20911 & 93.90 & 0.94 \\
& MMZ & 2479 & 55.21 & 0.54 \\
& MON & 3230 & 76.30 & 0.78 \\
& MYB & 5238 & 77.03 & 0.74\\
& NGB & 7967 & 72.31 & 0.74 \\
& NGS & 23628 & 93.03 & 0.92 \\
& PEB & 2196 & 76.84 & 0.77 \\
& PLM & 6137 & 91.22 & 0.93 \\
& PMO & 9546 & 83.37 & 0.86 \\
\bottomrule
\end{tabular}
\end{adjustbox}
\end{table}